\documentclass[runningheads]{llncs}

\usepackage[utf8]{inputenc}
\usepackage{xcolor}
\usepackage{amssymb}
\usepackage{amsmath}
\usepackage{amsfonts}
\usepackage{graphicx}
\usepackage{float}
\usepackage{xspace}
\usepackage{setspace}
\usepackage{textcomp}
\usepackage{xr-hyper}
\usepackage{hyperref}
\usepackage{csquotes}
\usepackage{comment}
\newcommand{\dnn}{\texttt{DNN}\xspace}
\newcommand{\probmodel}{\texttt{PM}\xspace}
\newcommand{\dataug}{\texttt{DA}\xspace}
\newcommand{\psr}{\texttt{PSR}\xspace}
\newcommand{\nnl}{\texttt{NLL}\xspace}
\newcommand{\ece}{\texttt{ECE}\xspace}
\newcommand{\bs}{\texttt{BS}\xspace}
\newcommand{\mce}{\texttt{MCE}\xspace}
\newcommand{\erm}{\texttt{ERM}\xspace}

\newcommand{\bdr}{\texttt{BDR}\xspace}
\newcommand{\arc}{\texttt{ARC}\xspace}
\newcommand{\vrm}{\texttt{VRM}\xspace}
\newcommand{\ce}{\texttt{CE}\xspace}
\newcommand{\acc}{\texttt{ACC}\xspace}
\newcommand{\mmce}{\texttt{MMCE}\xspace}
\newcommand{\version}[1]{\texttt{ARC\_V}_{#1}\xspace}

\usepackage{ifthen}
\newif\ifapxinsamepage

\apxinsamepagetrue

\ifapxinsamepage
{}
\else
{
\makeatletter
\newcommand*{\addFileDependency}[1]{% argument=file name and extension
  \typeout{(#1)}
  \@addtofilelist{#1}
  \IfFileExists{#1}{}{\typeout{No file #1.}}
}
\makeatother

\newcommand*{\myexternaldocument}[1]{%
    \externaldocument{#1}%
    \addFileDependency{#1.tex}%
    \addFileDependency{#1.aux}%
}
\myexternaldocument{main_appendix}
}
\fi

\begin{document}

\title{On Calibration of Mixup Training for Deep Neural Networks}
\author{Juan Maroñas\inst{1}\thanks{correspondence: jmaronas@prhlt.upv.es.}, Daniel Ramos\inst{2} \and Roberto Paredes\inst{1}}
\authorrunning{Maroñas et al.}
\titlerunning{On Calibration of Mixup Training for Deep Neural Networks}
\institute{PRHLT Research Center, Universitat Politècnica de València, Spain  \and AUDIAS - Audio, Data Intelligence and Speech, Universidad Autónoma de Madrid, Spain}

\maketitle              % typeset the header of the contribution
\begin{abstract}
Deep Neural Networks (\dnn) represent the state of the art in many tasks. However, due to their overparameterization, their generalization capabilities are in doubt and still a field under study. Consequently, \dnn can overfit and assign overconfident predictions -- effects that have been shown to affect the calibration of the confidences assigned to unseen data. Data Augmentation (\dataug) strategies have been proposed to regularize these models, being Mixup one of the most popular due to its ability to improve the accuracy, the uncertainty quantification and the calibration of \dnn. In this work however we argue and provide empirical evidence that, due to its fundamentals, Mixup does not necessarily improve calibration. Based on our observations we propose a new loss function that improves the calibration, and also sometimes the accuracy, of \dnn trained with this \dataug technique. Our loss is inspired by Bayes decision theory and introduces a new training framework for designing losses for probabilistic modelling. We provide state-of-the-art accuracy with consistent improvements in calibration performance.\footnote{Appendix and code are provided here: \href{https://github.com/jmaronas/calibration\_MixupDNN\_ARCLoss.pytorch.git}{GitHub link}}
\end{abstract}

\section{Introduction}
Deep Neural Networks (\dnn) are probabilistic models (\probmodel) that represent the state of the art in many tasks, either as end-to-end models \cite{pmlr-v97-tan19a}, or as part of complex decision systems \cite{Gulcehre:2017:ILM:3103639.3103741}. Many of the applications in which \dnn has widely overcome previous approaches require that the parameterized probability distributions are interpretable. This means that both the prediction, (for instance the class selected in a classification problem), and the probability assigned to that prediction, are important for the correct performance of the whole system. Examples of these applications are medical diagnosis \cite{medicaldiagnosis} or language recognition \cite{brummer06calibrationLanguage}. In all these problems, it is very different to decide towards an action with high probability, than doing it with a more moderated one. The ultimate consequences incurred in the decision process can be drastic if these probabilities are not reliable i.e., are not well-calibrated. In other words, our model is calibrated if the probabilities assigned reflect the uncertainty present in the data distribution. Moreover, our \probmodel  must be able to discriminate between the different classes, i.e. to separate them. Note that discrimination is inherent to the data distribution, which means that we cannot expect to separate our data if our data is not separable in its origins. Both good discrimination and a correct modelling of data uncertainty are mandatory to achieve optimal classification performance by the use of the Bayes Decision Rule (\bdr). 
    
The calibration and discrimination of a \probmodel can be improved by optimizing the expected value of a proper scoring rule (\psr), a scalar obtained by the addition of both quantities \cite{deGroot83forecasters}. For that reason, this optimization is not a guarantee of optimal calibration, as all the effort can be pushed into having better discriminative capabilities. This effect has been recently observed in the context of \dnn where \cite{DBLP:journals/corr/GuoPSW17} showed that although these models are typically trained by optimizing the Negative Log-Likelihood (\nnl), the calibration performance is compromised in the direction of over-confidence.  This means that even though the accuracy provided by these models on several benchmarks are among the best published, the probabilities assigned are ultimately extreme and badly calibrated. One should not be surprised about this generalization limitation, as many theories that study the generalization capabilities of probabilistic models, such as the VC dimension \cite{Vapnik1971OnTU} or the use of marginal likelihoods and Bayes rule for model selection \cite{mackayphd}, are instances of the Occam's Razor principle \cite{mackayphd}. For instance a recent work
 \cite{45820} has shown that \dnn can memorize the data input distribution and \cite{pmlr-v97-recht19a} has shown that many state of the art models overfit the test set.

For that reason, the community has been exploring different regularization techniques that can improve the generalization of these models, being Data Augmentation (\dataug) one of the gold standards. These techniques aim to increase the support on the input manifold, through transformations that are typically driven by expert knowledge, e.g. rotations or translations when the inputs are images. However, in many domains, it is not clear which kind of augmentations might be useful, which motivates the analysis of general-purpose \dataug techniques such as Mixup \cite{mixup}, whose fundamentals rely on empirical risk minimization (\erm) \cite{Vapnik1998}. However, both Mixup and human-driven \dataug techniques share a common issue: they are not designed by analyzing the properties of the input distribution and the intersection of these with the \probmodel; mainly because modern instances of these, such as \dnn, are difficult to interpret. For that reason, the selection and performance of \dataug techniques depend, basically, on cross-validation; but there is no principled way to establish if a particular \dataug technique might boost the performance of a particular application or not.

Motivated by the fundamentals and good performance of Mixup, a very recent work \cite{mixupcalibration} has studied how Mixup affects the uncertainty quantification and the calibration performance on \dnn. They show that Mixup improves the calibration, and they attribute this fact to the smoothness that Mixup induces in the decision regions learned by a \probmodel. Our work is built on top of this observation. We argue that the fundamentals of \erm and Mixup do not allow us to claim that learning smoother decision thresholds are a sufficient condition for having properly calibrated \probmodel, because this decision is not based on the uncertainty of our input distribution. This also extrapolates to other strategies that have shown good regularization in terms of accuracy, uncertainty quantification or calibration, such as label smoothing \cite{inception} or more recently \dataug techniques  \cite{unknown,hendrycks2019augmix}.

In this work, we first provide empirical evidence that Mixup can degrade calibration. Secondly, we propose a new loss function to correct this calibration degradation by encouraging the \probmodel to learn its discriminative capabilities, through the incorporation of a simple measure of data uncertainty. Thus, our loss function is inspired by how optimality is achieved in a \bdr scenario, and we claim that this has to be done to achieve reliable probability distributions. Note that learning to assign $\{0,1\}$ probabilities only makes sense if the input distribution does not present any kind of overlapping, which is something really hard to assess. For that reason, it should not be surprising that a modern \probmodel, such as a \dnn, can have undesirables effects such as memorization \cite{45820}  or overconfident-badly-calibrated probabilities \cite{DBLP:journals/corr/GuoPSW17} when forced to achieve this $\{0,1\}$ assignment, as it happens by learning through the categorical cross-entropy (\ce). Note that a modern \dnn, due to overparameterization, can successfully assign $\{0,1\}$ without any guarantees of generalization, and they typically rely on learning highly oscillating decision thresholds \cite{manifoldmixupicml}, which are also responsible for being vulnerable to adversarial attacks. The results of this work open new perspectives to design losses in this fashion, aiming at representing more sophisticated forms of data uncertainty.
\section{Related Work}
The first work that showed the badly calibrated probabilities of \dnn is found in \cite{DBLP:journals/corr/GuoPSW17}, where different classical calibration techniques are compared. The authors proposed Temperature Scaling. On top of this work \cite{maroas2019calibration} has shown how complex techniques can be employed for post-calibration if uncertainty is correctly incorporated, through the use of Bayesian Neural Networks. On the other hand, \cite{NIPS2019_9697,hendrycks2019using} has shown that using self-supervised learning and pre-trained models improves model robustness, uncertainty and calibration. Moreover, the same author has measured robustness against common perturbations \cite{hendrycks2019benchmarking}, and \cite{NIPS2019_9547} has measured the performance on calibration and uncertainty of several strategies under dataset shift. On the other hand, deep ensembles have also shown good performance for uncertainty quantification and calibration \cite{NIPS2017_7219}. Finally, on the side of \dataug strategies, \cite{mixupcalibration} measure the robustness and calibration of Mixup training and \cite{unknown} propose On-Manifold Adversarial Data Augmentation, which attempts to generate challenging examples by following an on-manifold adversarial attack path in the latent space of a generative model. Moreover, \cite{manifoldmixupicml} propose a similar technique to Mixup but on the hidden layers of a \dnn, with good results in robustness against perturbations. Finally, Augmix has been proposed in  \cite{hendrycks2019augmix} with good results in uncertainty quantification and robustness.
\section{Background}
In this section we describe calibration in the context of image classification and provide the fundamentals of Mixup before presenting our loss function in the next section. We are given $N$ pairs of observed i.i.d. samples $\mathcal{O}=\{(x_n,t_n)\}_{n=1}^N$ drawn from some unknown joint probability $P(x,t)$. We then learn a categorical posterior distribution $p_\theta(t = k|x)$ by means of a function $g_\theta$  that maps input images $x$ to class probabilities $\{k\}^C_{k=1}$  by maximum a posteriori. To make decisions we rely on \bdr and chose the action $\alpha_i$ that minimize Bayes Risk:
\begin{equation}
\begin{split}
    \mathtt{R}(\alpha_i|x) &= \underset{1\leq k \leq C}{\sum} \lambda_{ik} \cdot p(t = k|x)\\ 
    \alpha_i &= \underset{1\leq i \leq C}{\mbox{argmin}}\, \mathtt{R}(\alpha_i|x)  
    \label{equ:decision_rule}
\end{split}
\end{equation}
where $ \lambda_{ik}$ represents the loss incurred when taking the action $i$ if the ground truth is $k$. In this work we consider  equal losses $\lambda_{ik}=1,\lambda_{ii}=0\,\,\forall i,k$, which means that we choose the class with maximum posterior probability. This rule guarantees optimality when we plug in the data generating distribution \cite{DudaHartStork01}. In practice this distribution is substituted with the model $p_\theta(t|x)$ and thus, the lower the gap between the model and the data generating distribution, the closer we will be to an optimal decision.

In a classification scenario, we say that a model is calibrated if the confidences assigned by this model to a set of samples $X$ towards class $k$ are equal to the real proportion of samples in $X$ that the model assigns to this class. This means that to be calibrated, a model should assign confidences considering the proportion of samples assigned to each of the classes. Moreover, in addition to calibration, a model should also present a sharpened probability distribution, a property known as discrimination or refinement \cite{brummer06calibrationLanguage,deGroot83forecasters}. With this property, we guarantee that our model can discriminate between classes. Thus, both good calibration and discrimination
imply recovering how the data from the different classes is distributed or, in other
words, good calibration and discrimination imply recovering data uncertainty. By doing so, our model will be forced to match the data generating distribution and this will guarantee asymptotic optimality in the decisions to be taken.

Note that the goal of a \probmodel is to map any data distribution to a linear separable manifold. Thus, we can only achieve separability if: 1) the data is separable in its origins and 2) the model has enough capacity to do so. Thus, if 1) or 2) does not hold (which is something that we will not typically know), then it seems unreasonable to force the model to learn towards $\{0,1\}$ probabilities; and we should expect an overparameterized model to experiment different pathologies such as overfitting \cite{Vapnik1998}, memorization \cite{45820} or bad calibration \cite{DBLP:journals/corr/GuoPSW17}. A very illustrative example of this pathology is:  Why should we push probabilities towards $1.0$ in a 1-dimensional input generative Gaussian classifier if Gaussians have support over $\mathbb{R}$? Based on this observation a training loss in a modern \probmodel should somehow consider this inherent structure  (uncertainty) in the data to reliably target the underlying distribution, and avoid the great ability of \dnn to assign $\{0,1\}$ probabilities when we do not know if the distribution to be modelled is or can be linearly separated. This is the core idea of our proposed loss function, and we will further use it to justify why Mixup should not necessarily provide calibrated distributions. 

Mixup has its fundamentals in vicinal risk minimization (\vrm)\cite{VRM_NIPS}\footnote{For unfamiliar readers we provide a wider description in appendix \ref{mixup_explanation}.}, which is derived as a solution to the limitations present in \erm \cite{Vapnik1971OnTU,mixup,Vapnik1998}. Contrary to other vicinal distributions, Mixup assumes that the samples in the vicinity distribution do not belong to the same class. For that reason, it is defined as the expected value of a linear interpolation between two input samples and their corresponding labels \cite{mixup}. The interpolation is given by the coefficient $\gamma$, which is drawn from a beta distribution. An unbiased estimate of the empirical risk can be obtained by evaluating the average loss function on a set of samples drawn from this distribution as follows:
\begin{equation}
\begin{split}
    \gamma &\sim \mbox{Beta}(\beta,\beta)\\
    \Tilde{x} &= \gamma \cdot x_1 + (1-\gamma)\cdot x_2\\
    \Tilde{t} &= \gamma \cdot t_1 + (1-\gamma)\cdot t_2
\end{split}
\end{equation}
  As a consequence, training with Mixup ensures a linear-soft transition between the confidence assigned by a model in the different parts of the input space. However, this only ensures smoothness in the confidence assigned to different regions of the input space, reducing the overconfidence but without any guarantee of an improved calibration, because the uncertainty is not considered at all. Note that Mixup just relies on an assumption on how the samples in the vicinity are distributed but do not take into consideration the proportion of samples present, which is at the core of a proper calibration. 
 
 As a consequence, only if the data distribution presents a linear relation between their corresponding classes, one could expect the ultimate distribution to be calibrated when applying this technique. In the experimental section, we show that some models trained with Mixup do not necessarily improve the calibration, as recently noted in \cite{mixupcalibration}. In fact, we show that Mixup tends to worsen the calibration in many cases.

\section{Proposed Loss: Auto-Regularized-Confidence}
\label{sec:contribution}
As illustrated in previous sections, our objective is to benefit from the improved accuracy of Mixup, but providing better calibrated distributions. To do so, we introduce a new loss function, which is a weighted combination of our proposed loss, named Auto-Regularized-Confidence (\arc), and the categorical cross entropy (\ce). The \arc loss is inspired by the Expected Calibration Error (\ece)\cite{DBLP:journals/corr/GuoPSW17}\footnote{See appendix \ref{calibration_metrics} for a detailed description of calibration metrics.}. The idea, as argued in above sections, is to incorporate data uncertainty in the predictions. This is done by first partition the confidences $p$, assigned to a batch of samples $X$, into $M$ bins $B_i$; and match these confidences to the accuracy  $\mu_i$ in that bin, by means of any of these two variants:
\begin{equation}
\begin{split}
\version{1}&=\frac{1}{M}\sum^M_{i=1}\left[\left(\frac{1}{|B_i|}\underset{0\leq j \leq|B_i|}{\sum}p_{ij}\right) - \mu_i\right]^2\\
\version{2}&=\frac{1}{M}\sum^M_{i=1}\left[\frac{1}{|B_i|}\underset{0\leq j \leq|B_i|}{\sum}\left(p_{ij} - \mu_i\right)^2\right]
\end{split}
\end{equation}
 The difference lies in whether the average confidence ($\version{1}$) or the individual confidences ($\version{2}$)  are forced to match the accuracy. If we set $M=1$ then our loss function is computed over the entire batch. We make the accuracy $\mu_i$ a constant value so learning gradients only depend on the confidence assigned by the model. Our loss is combined with the \ce to avoid the local minimum in which the network parameterize a prior classifier (i.e., the one which assigns prior confidences to samples), as we found in our initial analysis. This is because a prior classifier is useless, but the trivial way of optimizing calibration. Thus the overall loss is given by:
\begin{equation}
        \mathtt{L}(\theta) = \frac{1}{N}\underset{n}{\sum} \ce(\theta,x_n,t_n) + \beta\cdot \text{\arc}(\theta,x_n,t_n)  
\end{equation}
where $\beta$ is a hyperparameter that controls the relative importance given to each of the losses and is established with a validation set. As mentioned, this new loss targets the uncertainty of the learned representation, through the accuracy.  The accuracy is used to summarize the proportion of samples from different classes that are being 
\enquote{mixed}. So it somehow represents how the representations that the model can learn are distributed. It is clear that the accuracy is a very simple statistical summary of the data uncertainty and it is let to future work the search for other quantifiers that could encode more useful information such as how samples are distributed in the input space.
 Consequently, we can expect that by evaluating the \ce loss on the Mixup image $\Tilde{x}$, and the \arc loss on the mixing images $x_1$ and $x_2$, one can benefit from the improved discrimination as learned by the \ce, but the ultimate confidences are assigned by how the classifier classifies samples $x_1,x_2$ from the generating distribution $p(x,t)$ and not those $\Tilde{x}$ virtually generated by Mixup. It is then clear that \arc incorporates data uncertainty, which will improve the model representation of the underlying distribution, and thus its calibration. To validate this procedure, in our work we experiment with variants that compute $\arc$ loss over $x_1$ and $x_2$; and we also compute $\arc$ loss over $\Tilde{x}$. In general, all datasets benefit more from the latter. A discussion is provided in the experimental section.

An additional analysis of this loss function is provided in appendix \ref{further_loss_analysis} and the experimental section. This includes the motivation beside experimenting with $\version{1}$ and $\version{2}$ and an analysis of why this loss might improve the accuracy, as we have found that some datasets improve this metric by applying the \arc loss.

Finally, we discuss one drawback of our proposal as being used as a general-purpose calibration tool. Note that, if applied on a \dnn that presents near $100\%$ accuracy on the training dataset (which is the case in many of the standard databases tested) then the $\arc$ loss will provide the same learning signal as the \ce, because it will for the average confidences to be $1.0$. This means that it will not work in datasets where the training error is overfitted, as in CIFAR100. To solve this, we experiment with the following variant. We take a validation split from the training dataset where the \dnn presents uncalibrated over-confidences.  Let say that this validation set presents an 80\% accuracy, with a $0.99$ average confidence. Thus, we use the validation set to compute the \arc loss while the training dataset is only used for the \ce.

\section{Experiments}
\label{experiment_section}
We perform several experiments that illustrate the main claims of this work. We show average results in the main work and provide specific results in Github, alongside code and details on loss hyperparameters (e.g if the model uses $\version{1}$ or $\version{2}$). We evaluate a collection of classical benchmarks for this task: CIFAR100, CIFAR10, SVHN; and we also evaluate our model on more realistic problems such as the ones provided by Caltech Birds and Standford Cars, which contain bigger and more realistic images. Due to computational restrictions, we did not evaluate our model on ImageNet. We experiment with state-of-the-art configurations of computer vision \dnn: Residual Networks, Wide Residual Networks and  Densely Connected Neural Networks. Moreover, for each variant, we evaluate several configurations and models with and without dropout. We use pre-trained models on ImageNet for Birds and Cars. We evaluate different calibration metrics, detailed in appendix \ref{calibration_metrics}. In the main work, we report the accuracy and \ece (with a partition of 15 bins) while the rest of the calibration metrics are reported in appendix \ref{additional_results}. We compare to a recent technique designed for implicitly calibrate a probabilistic \dnn named \mmce over their best performing approach \cite{pmlr-v80-kumar18a}.  More details provided in appendix \ref{experimental_details}.

For the sake of illustration, we provide average results of all the models in table \ref{average_results}, and for the best-performing model per task in table \ref{max_results}. First, as shown in rows B (Baseline) and B+M (Baseline+Mixup) in the tables, we see how Mixup degrades the calibration except in CIFAR100. By comparing with the results reported in \cite{mixupcalibration}, we can conclude that Mixup behaves particularly well in CIFAR100, probably because the intersection between classes can be explained through a linear relation. However, our tables demonstrate that this is not a general behaviour of Mixup as shown in the rest of datasets. It is surprising how Mixup degrades calibration in Birds and Cars, even though the \dnn used for these datasets are pre-trained models which have been shown to provide better calibrated distributions\cite{hendrycks2019using}. In general, our results contrast with those reported in  \cite{mixupcalibration} where they provide general improvement in calibration performance due to Mixup. We can explain this difference with the fact that different models are used. For instance, while they use a VGG-16 and a ResNet-34, we are using much deeper models, such as a ResNet-101 or a DenseNet-121. The difference can be connected to the observation in  \cite{DBLP:journals/corr/GuoPSW17} where they show that calibration is further degraded by deeper architectures. Moreover, we shall emphasize that our results on CIFAR10 are on the state-of-the-art ($\sim97\%$ \acc) and much better calibrated ($1.03$ top \ece and $1.62$ average \ece) than in \cite{mixupcalibration}, while they report a $2.00$ value of \ece. 

\begin{table}[!t]%
\centering
\caption{Table showing average accuracy and \ece in (\%) of all the models considered in this work}
\label{average_results}
{\renewcommand{\arraystretch}{1.21}
\resizebox{\textwidth}{!}{
\begin{tabular}{ccc|cc|cc|cc|cc}
& \multicolumn{2}{c}{CIFAR10} & \multicolumn{2}{c}{CIFAR100} & \multicolumn{2}{c}{SVHN}& \multicolumn{2}{c}{Birds} & \multicolumn{2}{c}{Cars} \\\cline{2-11}
  & \acc                  &  \ece   & \acc                 & \ece     & \acc                  & \ece   & \acc         &  \ece   & \acc        & \ece        \\\hline
Baseline (B)            & 94.76  & 3.41 & 77.21 & 11.57 & 96.32 & 1.90 & 78.51    & 2.39        & 86.74       & 2.06       \\
Baseline + Mixup  (B+M) & 96.01  & 4.35 & 80.04 & 3.71  & 96.41 & 5.00 & 79.63    & 14.22       & 86.67       & 18.13      \\\hline
\mmce (M)               & 94.24  & 2.17 & 72.68 & 3.71  & 96.28 & 1.78 & 78.78    & 1.95        & 86.83       & 2.23       \\
\mmce + Mixup  (M+M)    & 91.90  & 5.69 & 78.52 & 5.48  & 96.59 & 2.83 & 79.99    & 12.37       & 86.03       & 13.07      \\\hline
\arc   (A)              & 94.82  & 3.37 & 77.04 & 11.31 & 96.26 & 1.87 & 78.52    & 2.70        & 87.78       & 2.76       \\
\arc + Mixup (A+M)      & 95.90  & 1.62 & 79.84 & 2.42  & 96.02 & 2.17 & 79.74    & 4.95        & 89.63       & 2.84       \\\hline\hline
\end{tabular}}}
\end{table}

\begin{table}[!t]
\centering
\caption{Table showing the accuracy and \ece in (\%) of the best model per task and technique.}
\label{max_results}
{\renewcommand{\arraystretch}{1.21}
\resizebox{\textwidth}{!}{
\begin{tabular}{ccc|cc|cc|cc|cc}
                     & \multicolumn{2}{c}{CIFAR10}    & \multicolumn{2}{c}{CIFAR100}   & \multicolumn{2}{c}{SVHN}     & \multicolumn{2}{c}{Birds} & \multicolumn{2}{c}{Cars} \\\cline{2-11}
                     & \acc                  &  \ece   & \acc                 & \ece     & \acc                  & \ece   & \acc         &  \ece   & \acc        & \ece        \\\hline
Baseline   (B)          & 95.35 & 2.97 & 79.79 & 5.06 & 97.07 & 0.50 & 80.31 & 4.34  & 89.13 & 2.57  \\
Baseline + Mixup (B+M)    & 97.19 & 4.65 & 82.34 & 1.42 & 96.97 & 4.91 & 82.09 & 10.14 & 89.45 & 18.10  \\ \hline
\mmce   (M)             & 95.58 & 1.21 & 74.98 & 7.04 & 96.90 & 0.49 & 80.64 & 3.28  & 89.40 & 2.70  \\
\mmce + Mixup  (M+M)       & 97.02 & 1.11 & 81.31 & 4.46 & 97.17 & 3.69 & 82.41 & 10.93 & 88.47 & 11.56  \\\hline
\arc  (A)               & 95.99 & 2.01 & 80.77 & 4.73 & 97.08 & 0.37 & 80.32 & 4.44  & 90.09 & 1.92  \\
\arc + Mixup  (A+M)        & 97.09 & 1.03 & 82.02 & 0.98 & 96.82 & 2.20 & 82.45 & 1.28  & 91.13 & 2.40  \\\hline\hline
\end{tabular}}}
\end{table}

Analyzing our loss function, we see how it can correct the miscalibration introduced by Mixup training. In CIFAR10 and CIFAR100 A+M is the best performing approach. In SVHN we see that A+M corrects the calibration error introduced in B+M, but the approach behaves similar to the others. SVHN is a dataset that presents good calibration in many models over the test set, as noted also in \cite{DBLP:journals/corr/GuoPSW17,maroas2019calibration}. Finally, regarding Birds and Cars we see how our loss can highly correct the miscalibration introduced by Mixup. This means that our approach also performs well with pre-trained models on ImageNet. It should be noted that in this case, we do not achieve the same \ece error in Birds and Cars as with the baseline model. However, we have much better accuracy (over $3\%$ on average results in Cars). In fact, our work reports nearly state of the art accuracy in Cars using a Dense-Net, where the best performing reported model has an accuracy only two points above but using much more complex architectures such as efficient net \cite{pmlr-v97-tan19a} or inception \cite{inception}. On the other hand,  our method is better than the recently proposed \mmce \cite{pmlr-v80-kumar18a}. We found this method to be unstable in some cases, as some models saturated during training or tended to degrade the accuracy, as shown in the tables.

Regarding the parameterization of the loss function, we found that most of the times the best configuration of hyperparameters was obtained with $\version{1}$. This can be explained by the fact that \dnn typically learn invariant representations and thus, we avoid the pathological behaviour that $\version{1}$ can present, which is discussed in appendix \ref{further_loss_analysis}. Besides, we found that only in Birds and some CIFAR100 models, the \arc loss computed over the Mixup image $\Tilde{x}$ worked better than when computed over $x_1$ and $x_2$, even though this configuration also improved the calibration. Thus, as we claim in section \ref{sec:contribution}, it seems reasonable that a loss function that takes into account, separately, the underlying structure present in the data can provide better calibrated uncertainties.

Finally, by looking at the results of applying \arc loss over the Baseline model (A in the tables) we see that the improvements in calibration are not significant, or at least not as when combined with Mixup. We have already argued the reason in section \ref{sec:contribution}. We mentioned that a possible solution could be to apply the \arc loss on a separate validation set. Surprisingly, the \dnn  learns to minimize the $\arc$ loss by increasing the accuracy of this validation set rather than by relaxing the confidences assigned.

\section{Conclusions and Future Work}
This work has shown that Mixup does not ensure calibrated class distributions. The results and theory presented suggest that a similar analysis should be employed over different \dataug techniques, which is let for future work. We have also opened a new perspective to reduce overconfidence in \dnn. As we cannot control how a model might overfit the dataset to achieve high discriminative performance, a good practice is to auto-regularize the model to incorporate the uncertainty of the learned representations. This work has shown a way of doing this on Mixup training, reporting state-of-the-art results in accuracy and calibration. Future work is concerned with the exploration of new loss functions for this purpose.
\section{Acknowledgments}
We gratefully acknowledge the support of Nvidia-corporation through the donation of two Nvidia TITAN XP. The research leading to these results has received funding from the European Union through Programa Operativo del Fondo Europeo de Desarrollo Regional (FEDER) from ComunitatValencia(2014-2020) under project Sistemas de frabricaci{\'o}on inteligentes para la industria 4.0 (grant agreement IDIFEDER/2018/025). JM is supported by grant FPI-UPV under grant agreement 825111 Deep Health Project. DR and JM are supported by the Spanish National Ministry of Education through grant RTI2018-098091-B-I00.

\bibliographystyle{splncs04}

\ifapxinsamepage
{
%\bibliography{references_SSPR_with_appendix}

\clearpage

\begin{appendix}
\section{Mixup training}
\label{mixup_explanation}
In this appendix, we provide an extended explanation to the one in section \ref{sec:contribution}, regarding the reasons why Mixup should not necessarily provide calibrated distributions. Mixup has its fundamentals in vicinal risk minimization (\vrm) \cite{VRM_NIPS}, which is derived as a solution to the limitations present in \erm \cite{Vapnik1971OnTU,mixup,Vapnik1998}. The ideal learning signal in a frequentist paradigm should be provided by the gradient of the expected value of a loss function over the underlying probability density $P(x,t)$, 

\begin{equation}
    \mathcal{R}(\theta) = \int L(g_\theta(x),t)\,\mbox{d}P(x,t)
\end{equation}

\noindent which in practice is impossible as we only have access to an i.i.d. sample $\mathcal{O}$. As a consequence, we attempt to minimize the expected value of the loss function over the empirical distribution $P_d(x,t)$, a process known as empirical risk minimization (\erm). This distribution is given by Dirac delta distribution centred at the observed points:

\begin{equation}
    P_d(x,t) = \frac{1}{N} \underset{i}{\sum}\delta(x_i,t_i)
\end{equation}

Thus, \erm clearly lacks of support in many different parts of the input space, which makes this learning paradigm present some limitations such as over/under-fitting, memorization \cite{45820},  or sensitivity to adversarial examples \cite{42503}. \vrm is proposed to solve this lack of support in the input manifold. To achieve this goal, the Dirac Delta distribution is substituted with a \textit{vicinity} distribution, which aims at exploring different parts of the input space in the vicinity of the observed set $\mathcal{O}$. For instance, a vicinity distribution can be implemented as a Gaussian centred at each sample $x_i$. In practice, we then sample from this Gaussian distribution and recover an unbiased estimate of $\mathcal{R}(\theta)$ computed with this new set of generated samples, which is used in conjunction with stochastic gradient guided learning algorithms. Thus, any \dataug technique, such as Gaussian noise addition, can be understood under the \vrm paradigm.

The main motivation behind Mixup is that \dataug techniques assume that the samples in the vicinity distribution belong to the same class. For that reason, Mixup vicinity distribution is defined as the expected value of a linear interpolation between two input samples and their corresponding labels \cite{mixup}. This interpolation is parameterized by a coefficient $\gamma$ which is drawn from a beta distribution.  An unbiased estimate of $\mathcal{R}(\theta)$ can be obtained by evaluating the average loss function on a set of samples drawn from this distribution as follows:

\begin{equation}
\begin{split}
    \gamma &\sim \mbox{Beta}(\alpha,\alpha)\\
    x &= \gamma \cdot x_1 + (1-\gamma)\cdot x_2\\
    t &= \gamma \cdot t_1 + (1-\gamma)\cdot t_2
\end{split}
\end{equation}

As a consequence, training with Mixup smooths the predictions performed by a model in the intersection between samples from the unknown distribution $P(x,t)$. However, even if this might reduce high-oscillations in the predictions performed in these regions of the feature space, or smooth the ultimate confidence assigned to these regions, this only ensures that the model will be less overconfident, which does not necessarily mean that the ultimate probability distribution will be calibrated. This is because Mixup only ensures a linear-soft transition between the confidence assigned by the model in different parts of the input space. As a consequence, only if the data distribution presents a linear relation between their corresponding classes, one could expect the ultimate distribution to be calibrated. It is clear that Mixup interpolation does not consider the proportion of samples present in the input distribution, which is at the core of a proper calibration. In the experimental section we show that some models trained with Mixup do not necessarily improve the calibration, as recently noted \cite{mixupcalibration}. In fact, our results show that Mixup can highly degrade calibration in many cases.

\section{Measuring Calibration}
\label{calibration_metrics}
Calibration can be measured in different ways, each one with their own properties. While some metrics are directly \psr, such as the Brier score (\bs)\cite{deGroot83forecasters} or the logarithmic score (\nnl)\cite{deGroot83forecasters}, averaged over empirical samples; some others are merely measures of calibration, such as the expected calibration error (\ece) \cite{DBLP:journals/corr/GuoPSW17} and the maximum calibration error (\mce)\cite{DBLP:journals/corr/GuoPSW17}. Given a set of $M$ samples, each of these metrics can be computed in the following way:

\begin{equation}
\begin{split}
    \bs &= \frac{1}{M} \underset{n}{\sum} (g_\theta(x_n) - t_n)^2 \\
    \nnl &= -\frac{1}{M} \underset{n}{\sum} t_n\cdot \log [g_\theta(x_n)] \\
    \ece &=  \underset{j}{\sum}\frac{|B_j|}{M} |\mbox{acc}(B_j) - \mbox{conf}(B_j)|\\
    \mce &=  \underset{j}{\mbox{max}}\,\, |\mbox{acc}(B_j) - \mbox{conf}(B_j)|
\end{split}
\end{equation}

\noindent where the $[0,1]$ confidence range is equally divided into $j$ bins $B_j$. In each of these bins, the accuracy ($\mbox{acc}$) and the average confidence ($\mbox{conf}$) of the samples that lie in that particular bin are computed. In this case $t_n$ represents a one-hot vector for the class probability $k$.

Note, for instance, that the \nnl score highly penalizes important errors (i.e. extreme and wrong probabilities), but it is not able of separating which part of those errors are due to discrimination and which to calibration. This means that the \nnl will always be penalized under non-separable data manifolds even though we face the ideal situation in which the model has recovered the data generation distribution (thus, when it presents perfect calibration and has recovered data discrimination). In this situation, a perfect model will present $\ece=0$ because \ece is just a measure of calibration.

On the other hand, while $\ece$ is not sensible to high extreme errors made in only one sample (assigning 1.0 confidence towards an incorrect class), \nnl penalizes this error by assigning an infinity score. In this sense, \nnl is much more sensitive to strong overconfidence error than the rest of the calibration performance metrics and can be more useful in applications where overconfidence errors must be avoided in a very restrictive way.

\section{Further Loss Analysis}
\label{further_loss_analysis}
In this appendix, we provide additional analysis of the proposed loss function. First, note that while the \ce loss aims at pushing the probability of a given sample $x_i$ towards $1.0$ confidence of belonging to its associated class $t_i$, our loss encourages the model to auto-adjust its confidence depending on the accuracy of each batch of data being forwarded through the model. It is clear that this loss function and the \ce play different, and opposite, roles regarding the probabilistic information that the model should provide. For this reason, we might think that the combination of both losses could lead to a suboptimal result, as each of the losses pushes in opposite directions. In order words, both losses play a give-and-take game.  However, note that learning signals provided by the losses are somehow complementary. At the beginning of the learning processes, when the network is initialized at random, the network typically parameterizes a quasi-constant output distribution, and the accuracy provided by the model is near to that of a prior classifier. Thus, the learning signal provided by the \arc loss is negligible as compared to the one provided by the \ce. On the other hand, when the optimization of the \ce stalls, then the \arc loss plays its role by adjusting the ultimate confidences if they are uncalibrated.  This trade-off between \arc and \ce can be seen as a type of regularizer of the \ce by \arc, preventing \ce to reach discrimination without taking care of calibration. Our loss will not let the \ce push the probability towards extreme $1.0$ values.

Moreover, it should be noted that this cost presents other desirable properties that aim at improving regularization. First, consider a set of samples lying in the confidence range $[0.6,0.7]$. If the accuracy of these samples is located in this range then our loss function will encourage the model to adjust them to be as close as possible to the accuracy. Second, if the accuracy provided by the model has a value over this range, e.g $80\%$, then the model will raise these confidences to recover a calibrated model. It should be noted that in this case, our loss function will not change the accuracy as we are just pushing upwards the confidence of the samples which are originally correctly/incorrectly assigned, and thus the decision of which class should be assigned to each sample remains intact. Third, consider the same set of samples but with a provided accuracy of $40\%$. Our loss function will encourage these set of samples to reduce its confidences. It is clear that reducing this confidence has to be done at the cost of raising the confidence towards other classes. By doing this, we have a chance of changing the decision made by the model towards another class, thus helping to improve the discrimination of the model and consequently raising the accuracy. 

On the other hand, the idea of experimenting with the two variants of our loss named $\version{1}$ and $\version{2}$ is based on the following observation. The only difference between the two variants is whether we force the average confidence of a set of samples to match the accuracy, as performed by $\version{1}$, or we force each individual sample to match the accuracy, as done by $\version{2}$. $\version{2}$ is proposed to avoid solutions in which the set of confidences assigned by the model presents high variance. This will avoid solutions in which, for instance, the network present a $90\%$ accuracy on a set of samples, and the model assigns $0.8$ confidence to half of the samples and $1.0$ to the other half. In such a setting, the loss being minimized will be $0$, but the ultimate goal will not be achieved. The possibility of computing our loss over separate bins is incorporated to reduce this effect. However, in practice, we expect both losses to work, as the ideal behaviour of a good representation as learned by a model should be to map all the samples of a given class to the same (ideally linearly separable) representation. If this happens, the aforementioned variance on the confidence assigned by the model is reduced. 

\section{Additional Experimental Details}
\label{experimental_details}

\paragraph{Datasets} We choose datasets to evaluate our approach. We rely on classical benchmarks such as (number of classes into the brackets) CIFAR100 (100)\cite{cifar100}, CIFAR10 (10)\cite{cifar10}, SVHN (10)\cite{noauthororeditor}, and we also evaluate our model on more realistic problems such as the ones provided by  Caltech-Birds (200)\cite{WahCUB_200_2011}, Standford-Cars (196)\cite{KrauseStarkDengFei-Fei_3DRR2013}. These datasets are made up of bigger and more realistic images, and a padding preprocessing must be done. Due to computational restrictions, we did not evaluate our model on ImageNet.

\paragraph{Models.} We evaluate our model on several state-of-the-art configurations of computer vision neural networks, over the mentioned datasets: Residual Networks \cite{resnet}, Wide Residual Networks \cite{DBLP:journals/corr/ZagoruykoK16} and  Densely Connected Neural Networks \cite{DBLP:journals/corr/HuangLW16a}. Moreover, for each variant, we evaluate a model with and without Dropout. We find this interesting because a dropout model can be used to quantify uncertainties \cite{mcdropoutgal,NIPS2015_5666}. For the ResNet we add a Dropout layer after the whole network. We set the Dropout values according to the ones provided in the original works, or the model implementations, except for the ResNet where we use a $0.5$ Dropout rate. We use the pre-trained models on ImageNet provided by the PyTorch API for Birds and Cars datasets. On these pre-trained models, we add a Dropout layer at the end. Models are optimized with stochastic gradient descent with momentum and by placing a Gaussian prior over the parameters. The precision of this Gaussian prior is set accordingly to the provided implementations. For all the databases except Birds and Cars we use a learning rate starting from $0.1$. For Birds and Cars the initial learning rate is set to $0.01$. We use step learning rate scheduler that varies depending on the model. Additional details can be found in the code.

\paragraph{Data Augmentation Hyperparameters:} Regarding Mixup hyperparameters we used the ones provided in the original work. On the datasets where these techniques were not evaluated, we searched for the optimal value on a validation set. This hyperparameter is then fixed for the rest of the experiments carried out. More details on Github.

\paragraph{\arc Hyperparameters:}  Our loss hyperparameters: $\beta$, the number of bins and the type of cost used ($\text{V}_1$-$\text{V}_2$) were searched using a validation set with the ResNet-18 for all the models with and without Dropout. This is because we wanted to extract conclusions on a possible good configuration of our loss function and to do that we need to do a big battery of experiments (we trained more than 1000 Neural Networks to evaluate the loss); and this big experimental search came at the cost of computational restrictions.

As explained in the experimental section, this allows us to conclude that $\version{1}$ and bins $M=1$ are a reliable choice of the hyperparameters.  Our search include all the possible combinations of: loss $\version{1}$ and $\version{2}$; number of bins: $M=1$, $M=15$ and $M=\{5,15,30\}$ (for this one the loss is computed three times, one per each value of $M$, and the three losses are then averaged); and evaluation of the \arc loss over the Mixup image $\Tilde{x}$ or the separate images $x_1$ and $x_2$. This experiment was essential to validate our claim regarding data uncertainty and calibration, as exposed in section \ref{sec:contribution}. We run experiments over all these combinations, searching for the optimal $\beta$ value.  We select the value of $\beta$ that provides good accuracy with low calibration error. In some cases we found this hyperparameter to be $40$ times greater than the \ce loss, see details on Github. This enhances the beneficial influence that our loss function can have in several problems. 

Note that this way of searching for hyperparameters is not optimal.  In general, the extrapolated hyperparameter performed well in the rest of the models as detailed in the models on GitHub. However, sometimes, we experimented accuracy degradation in the training set. This is because a pathological solution of optimizing the \arc loss is by setting the parameters to output the data prior probability. This solution evaluates the \arc loss to $0$, but at the cost of parameterizing a useless prior classifier. As an example consider, for instance, that on the ResNet-18 we found that the optimal hyperparameter was $\beta=42$, but when training a DenseNet-121 this hyperparameter degraded the accuracy over the training set at the cost of providing perfect calibration. When this effect was observed we just picked the next hyperparameter that provided the next top performance over the ResNet-18; until the training accuracy was not degraded.

On the other hand, in CIFAR100 with Mixup we found this way of searching for the hyperparameter not to be as effective for the models without dropout. As provided in the specific results for each model trained on CIFAR100 in the tables provided in Github, we can see that all the models except ResNet-18 improve calibration when using Mixup. Thus, we cannot expect the hyperparameter to extrapolate as with other datasets. This was observed by training any of the deeper models with a validation set. To solve this, we simply perform a hyperparameter search over one of the deeper models in which Mixup showed great calibration performance, and use this parameter with the rest of the models. Due to computational limitations, we did not perform such an exhaustive search as we did with the ResNet-18, and just select a subset of the hyperparameters based on the previous wider analysis performed over the ResNet-18.

\subsection{Additional Results}
\label{additional_results}

We finally include additional results in our experiments. Table \ref{calib_metrics_average} show average results and table \ref{calib_metrics_indiv} show best performing model result for other calibration metrics. We can see that the results extrapolate from those in the experimental section, showing the improvement achieved by A+M.

Finally, table \ref{overfit_validation_set} shows the results of applying \arc loss only to a validation set that is uncalibrated. Surprisingly, the \dnn can increase the accuracy of this validation set without using the \ce, instead of relaxing the confidences. This shows the great ability of \dnn to overfit, and manifest the unpredictable behaviour of these models when used in probabilistic machine learning. This motivates the search of new losses that can encourage these powerful models to better represent the underlying distribution, and thus move them towards a better generalization, mandatory for critical applications.

\begin{table}[H]
\caption{This table shows different calibration metrics for average results. ACC in (\%), \mce in (\%), \bs$\times100$ and \nnl }
\label{calib_metrics_average}
\resizebox{\textwidth}{!}{
{\renewcommand{\arraystretch}{1.7}
\begin{tabular}{c||cccc||cccc||cccc||cccc||cccc|}
\multicolumn{1}{l}{} & \multicolumn{4}{c}{CIFAR10} & \multicolumn{4}{c}{CIFAR100} & \multicolumn{4}{c}{SVHN}   & \multicolumn{4}{c}{Birds}  & \multicolumn{4}{c}{Cars}   \\\hline\hline
Model               & \acc    & \mce  & \bs   & \nnl  & \acc    & \mce  & \bs   & \nnl & \acc    & \mce  & \bs   & \nnl & \acc    & \mce  & \bs   & \nnl & \acc    & \mce  & \bs   & \nnl  \\\hline
B                    & 94.76  & 2.16 & 0.86 & 0.23 & 77.21  & 5.20 & 0.35   & 1.08 & 96.32  & 1.86  & 0.62  & 0.17 & 78.51   & 0.58 & 0.17 & 1.04 & 86.74 & 0.56 & 0.10 & 0.52 \\
B+M                  & 96.01  & 2.88 & 0.65 & 0.18 & 80.04  & 0.67 & 0.29   & 0.79 & 96.41  & 2.76  & 0.63  & 0.18 & 79.63   & 1.49 & 0.18 & 1.11 & 86.67 & 1.81 & 0.13 & 0.71 \\
M                    & 94.24  & 1.06 & 0.89 & 0.21 & 72.68  & 0.61 & 0.38   & 0.98 & 96.28  & 1.12  & 0.61  & 0.17 & 78.78   & 0.46 & 0.17 & 1.02 & 86.83 & 0.59 & 0.10 & 0.52 \\
M+M                  & 91.90  & 2.73 & 1.33 & 0.32 & 78.52  & 1.18 & 0.31   & 0.86 & 96.59  & 1.46  & 0.59  & 0.16 & 79.99   & 1.23 & 0.17 & 1.07 & 86.03 & 1.28 & 0.12 & 0.67 \\
A                    & 94.85  & 2.13 & 0.85 & 0.23 & 77.04  & 4.78 & 0.35   & 1.05 & 96.26  & 1.16  & 0.62  & 0.17 & 78.52   & 0.65 & 0.17 & 1.04 & 87.78 & 1.32 & 0.10 & 0.51 \\
A+M                  & 95.90  & 0.78 & 0.67 & 0.17 & 79.84  & 0.54 & 0.29   & 0.80 & 96.02  & 1.11  & 0.64  & 0.16 & 79.74   & 0.65 & 0.15 & 0.82 & 89.63 & 1.70 & 0.08 & 0.44 \\\hline\hline
\end{tabular}}}
\end{table}

\begin{table}[!t]
\caption{This table shows different calibration metrics for the best model per task and technique. \acc in (\%), \mce in (\%), \bs$\times100$ and \nnl }
\label{calib_metrics_indiv}
\resizebox{\textwidth}{!}{
{\renewcommand{\arraystretch}{1.7}
\begin{tabular}{c||cccc||cccc||cccc||cccc||cccc|}
\multicolumn{1}{l}{} & \multicolumn{4}{c}{CIFAR10} & \multicolumn{4}{c}{CIFAR100} & \multicolumn{4}{c}{SVHN}   & \multicolumn{4}{c}{Birds}  & \multicolumn{4}{c}{Cars}   \\\hline\hline
Model                & \acc    & \mce  & \bs   & \nnl  & \acc    & \mce  & \bs   & \nnl & \acc    & \mce  & \bs   & \nnl & \acc    & \mce  & \bs   & \nnl & \acc    & \mce  & \bs   & \nnl  \\\hline
B                & 95.35 & 1.23 & 0.65 & 0.15 & 79.79 & 2.36 & 0.29 & 0.81 &  97.07 & 0.18 & 0.48 & 0.12 & 80.31 & 1.01 & 0.16 & 0.98 & 89.13 & 0.42 & 0.089 & 0.45  \\
B+M              & 97.19 & 3.11 & 0.47 & 0.14 & 82.34 & 0.46 & 0.26 & 0.70 &  96.97 & 2.55 & 0.53 & 0.16 & 82.09 & 1.12 & 0.15 & 0.97 & 89.45 & 1.84 & 0.110 & 0.61 \\
M                & 95.58 & 0.46 & 0.67 & 0.15 & 74.98 & 1.18 & 0.36 & 0.92 &  96.90 & 0.34 & 0.49 & 0.13 & 80.64 & 0.99 & 0.16 & 0.97 & 89.40 & 0.35 & 0.087 & 0.44 \\
M+M              & 97.02 & 0.73 & 0.45 & 0.11 & 81.31 & 0.70 & 0.28 & 0.74 &  97.17 & 1.36 & 0.50 & 0.14 & 82.41 & 1.05 & 0.15 & 0.96 & 88.47 & 1.14 & 0.105 & 0.59 \\
A                & 95.99 & 1.02 & 0.62 & 0.14 & 80.77 & 2.25 & 0.28 & 0.79 &  97.08 & 0.17 & 0.47 & 0.12 & 80.32 & 1.17 & 0.16 & 0.99 & 90.09 & 1.21 & 0.080 & 0.42 \\
A+M              & 97.09 & 0.39 & 0.48 & 0.12 & 82.02 & 0.31 & 0.26 & 0.72 &  96.82 & 1.75 & 0.51 & 0.14 & 82.45 & 0.34 & 0.13 & 0.69 & 91.13 & 0.91 & 0.078 & 0.42 \\
\hline\hline
\end{tabular}}}
\end{table}

\begin{table}[!t]
\caption{This table shows the results of applying the   \arc loss just to a validation set.}
\label{overfit_validation_set}
\resizebox{\textwidth}{!}{
{\renewcommand{\arraystretch}{1.7}
\begin{tabular}{c||cc|cc||cc|cc||cc|cc}
\multicolumn{1}{l}{} & \multicolumn{4}{c}{CIFAR100}                                                          & \multicolumn{4}{c}{CIFAR10}                                                           & \multicolumn{4}{c}{SVHN}                                                              \\\hline
\multicolumn{1}{c}{} & \multicolumn{2}{c}{validation}            & \multicolumn{2}{c}{test}                  & \multicolumn{2}{c}{validation}            & \multicolumn{2}{c}{test}                  & \multicolumn{2}{c}{validation}            & \multicolumn{2}{c}{test}                  \\\hline\hline
$\beta$              & \acc & \ece & \acc & \ece & \acc & \ece & \acc & \ece & \acc & \ece & \acc & \ece \\\hline
0.5                  & 82.74               & 4.32                & 77.97               & 9.48                & 97.28               & 1.29                & 95.29               & 2.82                & 98.50               & 1.21                & 96.45               & 2.32                \\
1.0                  & 86.46               & 4.29                & 78.74               & 10.65               & 97.92               & 0.88                & 94.92               & 3.39                & 98.86               & 0.54                & 96.48               & 2.33                \\
2.0                  & 91.26               & 2.17                & 79.49               & 9.19                & 99.08               & 0.27                & 95.25               & 3.05                & 99.08               & 0.33                & 96.54               & 2.37                \\
4.0                  & 94.30               & 1.61                & 79.61               & 8.86                & 99.74               & 0.21                & 95.60               & 2.63                & 99.24               & 0.24                & 96.50               & 2.33                \\
8.0                  & 96.26               & 1.08                & 78.85               & 9.73                & 99.84               & 0.18                & 95.58               & 2.72                & 99.32               & 0.18                & 96.67               & 2.13       \\\hline\hline        
\end{tabular}}}
\end{table}
\end{appendix}

}
\else
{
%\bibliography{references_SSPR}
\input{bbl_files/bbl_camera_ready.bbl}
}\fi

\end{document}